# Ethical Challenges in the Human-Robot Interaction Field

Julia Rosén
julia.rosen@his.se
University of Skövde
Skövde, Sweden

Jessica Lindblom
University of Skövde
Skövde, Sweden

Erik Billing
University of Skövde
Skövde, Sweden

Maurice Lamb
University of Skövde
Skövde, Sweden

## 1 INTRODUCTION

As the field of Human-Robot Interaction (HRI) matures, the frameworks, theories, models, methods, and tools that make up HRI are under continuous development [2]. One component often missing from this development is explicit discourse on the ethical impacts and outcomes of HRI research and application practices. While many ethical issues that arise in HRI correlates to the areas of Human-Computer Interaction (HCI), psychology, and sociology, the aims and practices of HRI research introduce unique ethical challenges [2, 7, 8, 20].

There are several main ethical frameworks which can be used to identify, classify, and judge ethical challenges. While we do not endorse any particular framework, it is useful to identify some features of a few relevant ethical frameworks. In all of these frameworks there is at least an actor (one who does an action to be deemed ethical), an action (an action with ethical implications), and a stakeholder (someone/thing affected by the action) [10]. Actors and stakeholders may include individuals, groups or communities, or the same individual in each category. In the following discussion, both utilitarian and deontological will be most relevant. In utilitarian frameworks, an action is ethically relevant when it increases/decreases some measured value of utility (benefit) to a stakeholder(s). In deontological frameworks, an action is ethically relevant if the actor must follow/violate a duty imposed on them as a member of society or a particular identity they hold. In the potential ethical challenges introduced below the most obvious issues arise due to risk of physical or psychological harm (a utilitarian concern) and the methodological centrality of deception (a violation of a potential deontological duty) [6, 13, 17].

The aim of this extended abstract is to introduce five examples of ethical issues in HRI that could have potential ethical implications, particularly on HRI participants. We consider these examples important to discuss in order to reach a consensus on how to handle them. Due to space limitations, this list is far from exhaustive and we hope that it can lead to a wider discussion that stimulates HRI researchers to think ethically. Previous work has shown a trend of underreporting ethical conduct in the HRI field [15]; in this extended abstract we consider some of the ethical issues that could arise in HRI research. We base the below discussion on three major codes of conduct that exist to guide researchers while performing research with human participants [1, 4, 19].

## 2 ETHICAL ISSUES

The first issue concerns situations where researchers *emulate emotions* in a robot in order to study the effect it has on participants. Many results from studies show that participants can indeed be affected by emotions in a robot [20]. A study by Gockley et al. [5] showed that people altered their behavior with a robot based on what facial expression the robot displayed (happy, sad, or neutral). Another study, by Niculescu et al. [12], showed that pitching the robot's voice, including changing of language cues (e.g. empathy and humor), resulted in changes in participant's overall feelings toward the robot. A common goal for these studies is to manipulate how participants react by changing the emotions a robot is exhibiting. As participants might not be aware of the robot's actual capabilities, this type of studies might constitute a form of deception [17]. Moreover, if we consider this a direct psychological effect on participants, one might consider the potential for psychological harm of a form typically requiring oversight by an ethical board in the context of psychology, and HCI. In this case, it is important to ensure that the participants are fully aware of how the study can affect them, e.g. by the use of informed consent and a debriefing session to inhibit the psychological effects [1].

The second issue concerns the *presentation form* of social robots. Thellman and Ziemke [18] showed that people's attitudes toward robots can easily be manipulated. Three versions of a questionnaire examining attitudes towards robots were constructed, each differing only in the human-likeness of a robot image on the cover. Despite the fact that none of the questions referred to the picture, the cover image affected participant responses. This finding highlights how easily people's attitudes toward robots, which still are relatively unfamiliar in most people's lives, might be manipulated. Manipulation is often seen as a potential harm requiring ethical oversight. Moreover, it is possible that researchers present robots in a way that misleads participants and the public. This deception may not be intentional, particularly given the strong role of popular fiction and non-fiction on preconceptions of robot abilities [9]. Notably, duties are often defined with respect to what is required for a community to exist. In order for HRI research to be valid it is critical to consider the ways that depictions of robots in media and commercial products impact perceptions. In particular, the HRI community should consider its role in shaping this media, both in the lab and through commercialization of ideas which are both important for the field.

The third issue concerns the *terminology* used when describing social robots, which may have an effect on people's views of social robots. McDaniel and Gong [11] highlighted that many authors in professional journals use personification language when writing about robots. For example, it is not uncommon to say that social robots have "eyes" and can "see." It may be obvious to HRI researchers that the robot cannot see in the way as the human visual perception system works, but it might not be as evident for participants. "When a robot is endowed with these kinds of physical human attributes, it is predictable that people will also credit it with a "brain" to control them. Rarely does one call the central processing unit, chip, or integrated circuits in an Apple computer a "brain"" [11, p. 179]. There are also instances when the terminology is misleading when describing certain roles that social robots may have. For example, one might be tempted to call social robots in health care, care robots. This may result in the expectation that robots are capable of caring. The terminology 'care robots' is therefore deceptive [14]. A more appropriate terminology could perhaps be 'robots for care'. It is well known that people ascribe agency to an objects [3], making these analogies potentially more likely to lead to unrealistic expectations — both inside and outside the lab.

The fourth issue concerns *data and privacy* when interacting with social robots. Both empirical research and technology development must grapple with data privacy issues. There are, however, some unique features when dealing with robots that interact with humans [16]. As mentioned above, many social robots are equipped with cameras, microphones, and other sensors. These sensors often fill multiple purposes. On the one hand, they may be used to directly record participants, but they may also be necessary for the robot to function properly. In the latter case, the researcher may unintentionally collect sensitive data in a way that is not obvious to either participants nor the researchers themselves. The terminology used to describe the robot's anthropomorphic features (e.g. eyes, voice, mouth) could also possibly contribute to the unawareness of the participants actually being recorded. Moreover as robots become integrated into household routines, the actions of the robots may unintentionally reveal data about household members. Even simple log data including information about when the robot is active/inactive may reveal potentially sensitive information, such as when the house is occupied or not.

The fifth issue concerns a specific experimental setup, most often referred to as *Wizard of Oz (WoZ)*. In this setup, a human is controlling the robot remotely, hidden from the participant [13]. The human control can range from small aspects like scripted dialogue and certain gestures, to full control over the robot. This is a common tool to investigate and analyze aspects of HRI that a robot cannot handle autonomously. While there is an ongoing discussion on what is ethically correct when using the WoZ set-up, there is no standardized practice available on how to handle it [6, 13]. Even when WoZ studies involves informing participants before the study, it is important to avoid unintentional deception through either terminology (e.g. anthropomorphism) or lack of clarity [17]. In the case of WoZ studies, it is critical that participants understand the actual capabilities of the robots as this will directly impact their expectations of robot capabilities moving forward. Robots that later do not meet expectations can result in a loss of trust. Thus, as with the some of the aforementioned issues, it may be a duty for HRI researchers to be transparent regarding robot capabilities.

## 3 CONCLUSIONS

With this work, we want to call attention to the ethical issues unique to HRI research and encourage ethical reflection for HRI practitioners. We advocate for proper ethical conduct, where ethical board approval, informed consent, data and privacy, deception, and debriefing is considered [15]. We believe this topic should be addressed with an interdisciplinary and systematic approach. The list of ethical issues mentioned in this extended abstract is not exhaustive and there are many more dimensions to this complex topic that are outside of the scope of this work. We consider several continuations in transforming research on e.g. robot ethics [20] into practices in the HRI field. We also acknowledge a common thread of ethical implications arising from the expectations of social robots, which deserves a wider discussion in future work.